\documentclass{article}
\usepackage{graphicx} 
\usepackage[accepted]{icml_fmwild}
\usepackage{microtype}
\usepackage{graphicx}
\usepackage{subfigure}
\usepackage{booktabs} 
\usepackage{hyperref}
\usepackage{amsmath}

\newcommand{\quotes}[1]{``#1''}

\begin{document}
\twocolumn[
\icmltitle{Bilingual Adaptation of Monolingual Foundation Models}

\icmlsetsymbol{equal}{*}

\begin{icmlauthorlist}
\icmlauthor{Gurpreet Gosal}{cs}
\icmlauthor{Yishi Xu}{cs}
\icmlauthor{Gokul Ramakrishnan}{cs}
\icmlauthor{Rituraj Joshi}{cs}
\icmlauthor{Avraham Sheinin}{cs}
\icmlauthor{Zhiming (Charles) Chen}{cs}
\icmlauthor{Biswajit Mishra}{cs}
\icmlauthor{Natalia Vassilieva}{cs}
\icmlauthor{Joel Hestness}{cs}
\icmlauthor{Neha Sengupta}{g42}
\icmlauthor{Sunil Kumar Sahu}{g42}
\icmlauthor{Bokang Jia}{g42}
\icmlauthor{Onkar Pandit}{g42}
\icmlauthor{Satheesh Katipomu}{g42}
\icmlauthor{Samta Kamboj}{g42}
\icmlauthor{Samujjwal Ghosh}{g42}
\icmlauthor{Rahul Pal}{g42}
\icmlauthor{Parvez Mullah}{g42}
\icmlauthor{Soundar Doraiswamy}{g42}
\icmlauthor{Mohamed El Karim Chami}{g42}
\icmlauthor{Preslav Nakov}{mbzuai}
\end{icmlauthorlist}

\icmlaffiliation{cs}{Cerebras Systems, Sunnyvale, CA}
\icmlaffiliation{g42}{G42, Abu Dhabi, UAE}
\icmlaffiliation{mbzuai}{Mohamed bin Zayed University of Artificial Intelligence, Abu Dhabi, UAE}
\icmlcorrespondingauthor{Gurpreet Gosal}{gurpreet.gosal@cerebras.net}
\icmlcorrespondingauthor{Avraham Sheinin}{avraham.sheinin@cerebras.net}

\vskip 0.3in
]
\printAffiliationsAndNotice{}
\begin{abstract}

We present an efficient method for adapting a monolingual Large Language Model (LLM) to another language, addressing challenges of catastrophic forgetting and tokenizer limitations. We focus this study on adapting Llama 2 to Arabic. Our two-stage approach begins with expanding the vocabulary and training only the embeddings matrix, followed by full model continual pre-training on a bilingual corpus. By continually pre-training on a mix of Arabic and English corpora, the model retains its proficiency in English while acquiring capabilities in Arabic. Our approach results in significant  improvements in Arabic and slight enhancements in English, demonstrating cost-effective cross-lingual transfer. We perform ablations on embedding initialization techniques, data mix ratios, and learning rates and release a detailed training recipe. To demonstrate generalizability of this approach we also adapted Llama 3 8B to Arabic and Llama 2 13B to Hindi.

\end{abstract}

\section{Introduction}

There has been a rapid advancement in open source English-dominant foundation language models like Llama 2 \cite{touvron2023llama}, Mistral 7B \cite{jiang2023mistral}, and Llama 3, primarily trained on extensive English corpora with minimal inclusion of non-English languages. 
To create models proficient in low-resource languages, two approaches can be taken: training a bilingual or multilingual model from scratch or adapting an existing strong English-dominant model to the target language.
While bilingual and monolingual models trained from scratch, like Jais \cite{sengupta2023jais} and Bloom \cite{workshop2023bloom}, have shown promise in non-English capabilities, they are expensive to train and have inferior capabilities in English.
Adapting strong English-dominant models to new languages also pose challenges, such as catastrophic forgetting of English capabilities, inefficiencies of English-dominant tokenizers, and the need for hyperparameter adjustments. \cite{fujii2024swallow, luo2024empirical,  FRENCH1999128, huang2024acegpt}.  
Here we address the challenges of the model adaptation approach.

Knowledge, reasoning and truthfulness capabilities  of Large Language Models (LLMs) are transferable across languages  \cite{yang2024large, sengupta2023jais}. This gives us the basis and motivation to explore efficient methods for cross-lingual transfer from English to Arabic through continual pre-training  of a monolingual English-dominant LLM without degradation of English capabilities.
Several recent works demonstrate cross-lingual transfer of foundation models  \cite{de-vries-nissim-2021-good, marchisio-etal-2023-mini, csaki2023efficiently, zhao2024Llama, huang2024acegpt, da-dalt-etal-2024-flor-effectiveness}, yet they lack comprehensive analysis of hyperparameter tuning, tokenizer, data mix selections, and  the impact of different model sizes.

We study the following aspects of cross-lingual adaptation. 

\textbf{Vocabulary extension}
We find that adapting a pre-trained model to a new language requires expanding the vocabulary, along with employing the methods below to maintain the model's original capabilities while acquiring new linguistic skills. We determine the optimal extension ratio of the original vocabulary through experimentation.

\textbf{Embedding alignment}
We find that ensuring alignment between the embeddings of the original and newly added vocabulary tokens is vital. We explore three techniques for initializing newly added token embeddings. We follow with embedding-only pre-training, which further aligns the embedding scale and orientation for original and new tokens.

\textbf{Continual pre-training}
Following the embedding alignment, we continually pre-train the model. We conduct experiments at the 7B model scale to assess various English-Arabic mix ratios and learning rates. We leverage the insights obtained from these experiments to perform cross-lingual adaptation to Arabic with Llama 2 13B and Llama 2 70B models.


Careful empirical study of vocabulary extension, embedding alignment, data mixtures and hyperparameter tuning, results in a recipe for language adaptation with significant performance improvements in Arabic and, uniquely, enhancements in English on Llama 2 models. To demonstrate the generalizability of this recipe to other languages and models, we adapt Llama 3 8B to Arabic and  Llama 2 13B to Hindi.



\section{Pre-training Datasets}
We use the AraV5 Arabic dataset, which includes documents from various sources such as web pages, Wikipedia, Arabic news outlets, books and social media. It also includes high-quality English corpora, Books and Wikipedia, translated to Arabic. Curated by \cite{sengupta2023jais}, it was used for pre-training the Jais series of Arabic foundation models. Prior domain adaptation studies have emphasized the importance of using "replay" data, which aligns with the pre-training data domain, to preserve the foundational model's knowledge and reasoning capabilities, as proposed in works by \cite{gupta2023continual,chen2023meditron70b,azerbayev2024llemma}. We use the Pile corpus, comprising data from 22 diverse sources including ArXiv, Wikipedia, PubmedCentral, CommonCrawl, OpenWebText, and Github \cite{gao2020pile}. For Llama 2 7B adaptation, we utilize 20B tokens from AraV5, whereas for Llama 2-13B and 70B, we utilize the entire AraV5 corpus, totaling 140B tokens. For Hindi adaptation of Llama 2 13B we use a Hindi corpus of 65B tokens curated from diverse resources and mix it with Pile.


\section{Methodology}
\label{method}


\subsection{Vocabulary Extension}

The first step in adapting a monolingual foundation model for multilingual use is to construct a balanced vocabulary that includes all target languages. Recent state-of-the-art models such as Llama 2\cite{touvron2023llama} use byte pair encoding (BPE) \cite{sennrich-etal-2016-neural} tokenizers, primarily  trained on English data. These tokenizers often split non-English words into characters or bytes, creating a significant imbalance among languages. Fertility, which measures the average number of subwords produced by a single word upon tokenization, can be used to quantify this imbalance.

This imbalance introduces inefficiency in pre-training, fine-tuning and inference. Table \ref{tab:vocab_extension} shows that the Llama 2 tokenizer needs as many as $4$ times the number of tokens to represent the same Arabic text as Jais’ Arabic-English bilingual tokenizer (MLV2) \cite{sengupta2023jais}. Balanced multilingual tokenizer with low fertility in all languages offers three main advantages \cite{petrov2023language}: 1) lower training and inference cost; 2) reduced latency during inference; 3) longer context windows. Models trained with low fertility tokenizers tend to perform well on downstream tasks as shown in \cite{ahuja-etal-2023-mega}

We experiment with two methods, vocabulary replacement and vocabulary extension, to create balanced tokenizers for English and Arabic. Vocabulary replacement implies maintaining the base vocabulary and replacing its least frequent tokens with the most frequent Arabic tokens. Vocabulary extension adds the most frequent Arabic tokens, increasing the vocabulary size. In both methods, we ensure that the newly introduced tokens are not present in the original vocabulary. For both methods, we determine the optimal number of new tokens to create a balanced multilingual vocabulary. Using Arabic tokens from the MLV2 vocabulary, we create two candidate tokenizers and perform intrinsic and extrinsic evaluations following \cite{ali2024tokenizer}. 

For intrinsic evaluation, we use fertility score to measure the efficiency of the tokenization process. We define fertility as \begin{math} f = \frac{S}{W} \end{math}, where \textit{S} is the total number of tokens in the tokenized text and \textit{W} is the number of words in the raw text. 
Subsets of validation sets of Pile and AraV5 are used to calculate the English and Arabic fertility, respectively. Table \ref{tab:vocab_extension} shows the intrinsic evaluations of two tokenizers, i) \textbf{\textit{Llama 2-replace30}},  and ii) \textbf{\textit{Llama 2-extend100}}. \textit{Llama 2-replace30} replaces 30\% of the base Llama 2 tokens while \textit{Llama 2-extend100} extends the Llama 2 vocabulary by 100\%. \textit{Llama 2-extend100} reduces the fertility of Llama 2’s tokenizer by 72.17\% while maintaining the fertility in English. It also reaches a fertility in Arabic comparable to MLV2. 

We perform extrinsic evaluation by continually training Llama 2 7B on a mixture of AraV5 and Pile, and monitoring the AraV5 validation loss. 
For a fair comparison of the tokenizers, we fix the raw text bytes at $67$ GB for Pile and $345$ GB for AraV5. Using \textit{Llama 2-extend100} as the candidate tokenizer and \textit{Llama 2} as the baseline, we tokenize the raw corpora and continually pre-train Llama 2 7B. Although the base Llama 2 tokenizer achieves a lower AraV5 validation loss compared to \textit{Llama 2-extend100} (see Table \ref{tab:vocab_extension}), it is trained on significantly more Arabic tokens due to its $\approx 3.5$ times higher fertility in Arabic. In an iso-token comparison, where the number of AraV5 tokens is fixed, \textit{Llama 2-extend100} outperforms base Llama 2 tokenizer by $\approx 2\%$. Considering both the intrinsic and extrinsic evaluations, we select \textit{Llama 2-extend100}. We correct all losses to align with the Llama 2 tokenizer (see \ref{tokenizer_appndx} and \ref{cec}).
We performed a similar analysis for Llama 3 8B adaptation to Arabic. Llama 3 8B has a four times larger vocabulary than Llama 2 and possesses superior multilingual capabilities including in Arabic. We found that extending it by 20\% brought the Arabic fertility within 10\% of the MLV2 tokenizer's fertility rate for Arabic. For Hindi adaptation of Llama 2 we extended the base vocabulary by 100\% just as for Llama 2 Arabic adaptation. 

\begin{table*}[h!]
    \centering
    \small
    \begin{tabular}{ c | c c | c c}
        \hline
        \hline
                                    & \textbf{Llama 2}  &  \textbf{MLV2 (Jais)}  & \textbf{Llama 2-replace30}  & \textbf{Llama 2-extend100} \\ 
        \hline
        vocab size                  & 32,000            & 84,992                 & 32,000                      & 64,000 \\ 
        \hline
        En. Fertility               & 1.86              & 1.62 (-12.63\%)        & 1.92 (+3.73\%)              & \textbf{1.85 (-0.03\%)} \\
        Ar. Fertility               & 5.06              & 1.29 (-74.55\%)        & 1.66 (-67.26\%)             & \textbf{1.41 (-72.17\%)}\\ 
        Ar val. loss                & \textbf{0.6371}   & -                      & 0.6440                      & 0.6539 \\
        IsoToken Ar val. loss       &  0.6668           & -                      & -                           & \textbf{0.6539} \\ \hline     
    \end{tabular}
    \caption{Tokenizer intrinsic and extrinsic evaluation. We see that MLV2 tokenizer reduces the fertility in Arabic by 75.55\%, \textit{Llama 2-replace30} by 62.26\%,  \textit{Llama 2-extend100} by 72.17\% compared to Llama 2 tokenizer. }
    \label{tab:vocab_extension}
\end{table*}

\subsection{Embedding initialization}
\label{emb_init}
For \textit{Llama-extend100}, we add $32000$ new Arabic tokens to the Llama 2 vocabulary, expanding the embedding and unembedding layers as \begin{math} [32000, d] \rightarrow [64000, d]  \end{math} where \textit{d} is the hidden size of the transformer decoder block. Our studies showed that the choice of embeddings initialization for newly added tokens is critical such that tokens which represent similar concepts in two languages are closer in the embeddings latent space. Simple methods such as Xavier \cite{pmlr-v9-glorot10a} or Kaiming \cite{kaiming} initialization, or using the mean of the embedding vectors of Llama 2 tokens, do not yield satisfactory results. Therefore, we explore alternative methods, described below, which demonstrated superior performances.

\textbf{Similarity-Based Token Embedding Initialization} \\
This method is inspired by the approach proposed in \cite{minixhofer-etal-2022-wechsel}. For each new token, we identify the top-\( k \) similar tokens using cosine similarity in the base vocabulary, using an external embedding. We use OpenAI's \texttt{text-embedding-3-large} embeddings \cite{kusupati2024matryoshka} for their superior quality and multilingual performance. Using cosine similarity, we find top-\( k \) similar tokens for the new token and initialize the new token embeddings by taking the weighted average of base embeddings of these similar tokens. After experimenting with different values for the \( k \), we achieve the best results with \( k=5 \). This initialization method was used for embeddings and unembeddings layers of Llama 2 13B adapted to Hindi.   

\textbf{Embedding Space Transformation} \\
In this initialization method, we leverage the pre-trained embedding vectors of Jais-30B \cite{sengupta2023jais30b}. We use $21377$ embedding vectors corresponding to tokens present in the intersection of the Llama 2 and Jais vocabularies to transform the Jais embeddings of the added tokens to the Llama 2 embedding space. Let \( \mathbf{E_{Jais}} \) and \( \mathbf{E_{Llama 2}} \) to denote the embedding matrices of the overlapping tokens of Jais and Llama 2,   
\begin{math} 
  \mathbf{E_{Jais}} \in R^{21377\times 7168}
\end{math} 
and 
\begin{math} 
  \mathbf{E_{Llama 2}} \in R^{21377\times4096}
\end{math}. 
We find a linear transformation to project \( \mathbf{E_{Jais}} \) to \( \mathbf{E_{Llama 2}} \)'s space by solving for \( W \) and \( b \) using the least squares method,   
\begin{math}
    W \mathbf{E_{Jais}} + b = \mathbf{E_{Llama 2}}
\end{math}.
 
We find  \( W \) and \( b \) such that the Euclidean 
\( \ell_2 \) norm \( \| W \mathbf{E_{Jais}} + b -  \mathbf{E_{Llama 2}} \|_2 \) is minimized. 
The parameters \( W \) and \( b \) are then used to project added tokens into the Llama 2 embedding space. This method performs better than similarity-based initialization (see  \ref{emb_init_appndx}).

\subsection{Embedding-only pre-training}
Even with Embedding Space Transformation initalization, the scale and the orientation of English and resulting Arabic embeddings are not aligned. Following \cite{de-vries-nissim-2021-good}, we do embedding-only pre-training using 15 billion tokens of AraV5 and Pile, mixed in 9:1 ratio. During this stage, gradient updates are applied to the embedding and the unembedding layers only, while keeping the other layers frozen. In our experiments, this method resulted in up to 2\% improvement in upstream loss for Arabic. 

\subsection{Hyper-parameter tuning} \label{hyperparameter_tuning}
  Hyperparameter sweep is important to determine best hyper-parameters such as learning rate (lr), warm-up schedule and batch size (bs). We use a linear warm-up for 1\% of the total steps followed by cosine decay \cite{cos-dcay2017} to 1/10th of the peak learning rate. We compare batch sizes of 4M tokens and 6M tokens but don't see a significant difference in upstream losses. We pick 4M tokens as the final batch size. We setup a learning rate sweep taking three different learning rates in different ranges. Let $lr_{peak}$ be the peak Llama 2 learning rate which is $3e\text{-}4$, we \quotes{re-warm} the learning rate to i) $lr_{peak}$, ii) $lr_{peak}/2$, and iii) $lr_{peak}/4$. These experiments use $14$ billion AraV5 tokens mixed with Pile in $1:1$, $3:1$ and $9:1$ ratios.  Across all ratios we find that $lr_{peak}$ performs the best as shown in table \ref{tab:hyper_parameter}. We used same hyperparameters for Hindi adaptation of Llama 2 13B. 
  For Llama 3 8B we performed a similar sweep and found that for stable adaptation to new domain. a smaller learning rate of $7.5e\text{-}5$, a larger batch size of 6M tokens is required.  
  \begin{table*}[h!]
    \centering
    \small
    \begin{tabular}{c|c c c c c c}
        \hline
        \hline
         \textbf{lr}     & \textbf{Mix En:Ar}   &  \textbf{Total tokens}  &  \textbf{AraV5 Tokens}   & \textbf{Pile Tokens}   &   \textbf{Pile eval loss} & \textbf{AraV5 eval loss} \\ \hline
        7.5e-5       &  1:9    &  15.6B        &  14B          &  1.56B        & 1.5397   & 2.5965 \\
        1.5e-4       &  1:9    &  15.6B        &  14B          &  1.56B        & 1.5396   & 2.3741 \\
        3e-4       &  1:9      &  15.6B          &  14B          &  1.56B        &  1.546  & \textbf{2.28} \\
        \hline
        7.5e-5       &  1:3    &  18.67B        &  14B          &  4.67B        &  1.5225 & 2.556 \\
        1.5e-4       &  1:3    &  18.67B       &  14B          &  4.67B        &   1.5205 & 2.3628 \\
        3e-4       &  1:3      &  18.67B         &  14B          & 4.67B        &  1.5234  &  \textbf{2.24}\\
        \hline
        7.5e-5       &  1:1    &  28B        &  14B          &  14B        &  1.5044  & 2.5135 \\
        1.5e-4       &  1:1    &  28B        &  14B          &  14B        &  1.5007  & 2.349 \\
        3e-4       &  1:1      &  28B          &  14B          &  14B        &  1.5093  & \textbf{2.2135} \\
        \hline
        
    \end{tabular}
    \caption{Llama 2-7B learning rate ablation experiments with different English to Arabic data mixtures. Data was tokenized with \textit{Llama 2-extend100} tokenizer and embeddings were initialized with subword-mean \ref{emb_init} approach. Base Llama 2-7B's Pile and AraV5 validation loss is  $1.5466$ and $2.95$, respectively. } 
    \label{tab:hyper_parameter}
\end{table*}
\subsection{Data mixture}
\label{data_mix_sub}




Domain adaptation involves continual pre-training a foundation model on new data not seen during the pre-training. When this new domain data is out-of-distribution, it can cause significant forgetting of prior capabilities which is  referred to as \textit{stability gap} \cite{guo2024cpt}. Adding a small proportion of replay data, which is closer in distribution to the pre-training data, can mitigate the forgetting. We conduct exhaustive experiments to find a minimum proportion of Pile data that should be mixed with AraV5 to mitigate forgetting. Table \ref{tab:hyper_parameter} shows results from the experiments with different data mixes. We found that mixing $1$ part English with $9$  parts Arabic ($1:9$ En:Ar) is sufficient to mitigate forgetting. We also don't see any forgetting in downstream evaluation as discussed in section \ref{downstream}. Interestingly, increasing the amount of English data while keeping Arabic tokens constant improves Arabic performance, indicating cross-lingual capability transfer. We leave the exploration of cross-lingual transfer to future work.

For Arabic adaptation of Llama 3 8B we found that a higher proportion of replay data is necessary.. Therefore, $1:1$ English to Arabic dataset mixture was used where we saw cross-lingual capability transfer between the two languages similar to Llama 2 adaptation. For replay data  we needed a mix of textbooks, math, coding and reasoning datasets mitigate forgetting. 

\section{Results}

\begin{table*}[h!]
    \centering
    \small
    \resizebox{\textwidth}{!}{\begin{tabular}{l|l l l|l}
    \hline
    \hline
     Model & Knowledge Average & Commonsense Reasoning Average & Misinformation, Bias Average & Overall Average \\ \hline
        llama 2-70b adapted & 38.4\% & 52.1\% & 51.4\% & 49.2\% \\ 
        llama 2-13b adapted & 34.1\% & 49.2\% & 48.8\% & 46.1\% \\ 
        llama 2-7b adapted & 33.8\% & 46.1\% & 49.1\% & 43.5\% \\ \hline
        llama 2-70b base & 31.4\% & 42.8\% & 48.9\% & 41.8\% \\ 
        llama 2-13b base & 29.4\% & 40.3\% & 47.7\% & 39.6\% \\ 
        llama 2-7b base & 27.3\% & 39.3\% & 47.5\% & 38.5\% \\ \hline
        Acegpt-13b & 32.3\% & 45.4\% & 50.8\% & 43.8\% \\ \hline
        Jais 30b & 38.0\% & 52.1\% & 51.2\% & 49.1\% \\ 
    \end{tabular}}
    \caption{Arabic summarized comparisons between: (1) Llama 2 models pre and post adaptation; (2) other notable English-Arabic bilingual models: Jais 30b \cite{sengupta2023jais30b} and AceGPT \cite{huang2024acegpt}}
    \label{tab:Arabic_downstream}
\end{table*}

\begin{table*}[h!]
    \centering
    \small
    \resizebox{\textwidth}{!}{\begin{tabular}{l|l l l|l}
    \hline
    \hline
         Model & Knowledge Average & Commonsense Reasoning Average & Misinformation, Bias Average & Overall Average \\ \hline
        llama 2-70b adapted & 45.9\% & 66.6\% & 56.0\% & 60.9\% \\ 
        llama 2-13b adapted & 38.4\% & 62.8\% & 49.3\% & 55.9\% \\ 
        llama 2-7b adapted & 36.2\% & 57.8\% & 51.6\% & 52.3\% \\ \hline
        llama 2-70b base & 48.8\% & 64.4\% & 57.1\% & 60.2\% \\ 
        llama 2-13b base & 37.9\% & 60.8\% & 53.7\% & 55.3\% \\ 
        llama 2-7b base & 36.1\% & 58.9\% & 55.4\% & 54.1\% \\ \hline
        Acegpt-13b & 37.2\% & 62.0\% & 56.6\% & 56.5\% \\ \hline
        Jais 30b & 41.3\% & 64.6\% & 56.3\% & 58.8\% \\ 
    \end{tabular}}
    \caption{English summarized comparisons between: (1) Llama 2 models pre and post adaptation; (2) other notable English-Arabic bilingual models: Jais 30b \cite{sengupta2023jais30b} and AceGPT \cite{huang2024acegpt}}
    \label{tab:English_downstream}
\end{table*}

\begin{table*}[h!]
    \centering
    \small
    \resizebox{\textwidth}{!}{\begin{tabular}{l|l l l|l}
    \hline
    \hline
         Model &Knowledge Average & Commonsense Reasoning Average & Misinformation, Bias Average & Overall Average \\ \hline
        llama 3-8b base  & 33.0\% & 45.2\% & 47.3\% & 43.2\% \\ 
        llama 3-8b adapted  &38.8\% &  49.2\% &53.3\% & 47.9\% \\ \hline
    \end{tabular}}
    \caption{Arabic summarized comparisons between Llama 3 8B model pre and post adaptation where we see  a 4.5\% average improvement across all domains}
    \label{tab:Llama3_downstream_Arabic}
\end{table*}

\begin{table*}[h!]
    \centering
    \small
    \resizebox{\textwidth}{!}{\begin{tabular}{l|l l l|l}
    \hline
    \hline
         Model &Knowledge Average & Commonsense Reasoning Average & Misinformation, Bias Average & Overall Average \\ \hline
        llama 3-8b base  & 51.1\% &63.7\% & 53.2\% & 59.5\% \\ 
        llama 3-8b adapted  &49.9\% & 64.0\% & 58.9\% & 60.5\% \\ \hline
    \end{tabular}}
    \caption{English summarized comparisons between Llama 3 8B model pre and post adaptation where we see  a 1\% average improvement across all domains. There is a 1\% drop in knowledge tasks which can be attributed to stability gap during domain adaptation}
    \label{tab:Llama3_downstream_En}
\end{table*}

\begin{table*}[htbp!]
    \centering
    \small
    
    \begin{tabular}{l|c|c}
    \hline
    \hline
         Model & XNLI Hindi & English downstream average \\ \hline
        llama 2-13b base  & 37,44\% & 57.1\%\\ 
        llama 2-13b adapted  & 45.68\%  & 55.3\% \\ \hline
    \end{tabular}
    \caption{Downstream performance comparison between Llama 2 13B pre and post adaptation on i) XNLI-Hindi, ii) English downstream task average of World Knowledge tasks, Commonsense reasoning tasks, and Misinformation and Bias tasks.}
    \label{tab:Llama2_downstream_Hi}
\end{table*}

\label{downstream}
Using the methodology described in section \ref{method} we adapt Llama 2 7B, 13B and 70B models to Arabic. We use linear warm up of the learning rate to $lr_{peak}$ for the first $1$\% of the tokens followed by cosine decay to $1/10$th of the $lr_{peak}$. For 7B and 13B models, we use $1:1$ En:Ar mix as we show in section \ref{data_mix_sub} that higher proportion of English also improves Arabic performance. For Llama 2 70B we use $1:9$ En:Ar mix to reduce compute time. Llama 2 7B adaptation uses $20$ billion tokens each from AraV5 and Pile. Llama 2-13B and Llama 2 70B use all $140$ billion tokens from AraV5 and $140$ billion and $15.56$ billion tokens from Pile, respectively. Tables \ref{tab:Arabic_downstream} and \ref{tab:English_downstream} show the $0$-shot downstream performance of the resulting models against the base models and other Arabic models like Jais \cite{sengupta2023jais30b} and AceGPT \cite{huang2024acegpt}. For Arabic evaluations we translated the English downstream tasks datasets using a similar approach as in \cite{sengupta2023jais}.

We evaluated the models on the \textbf{World Knowledge tasks}  MMLU \cite{hendrycks2021measuring} RACE \cite{lai-etal-2017-race} and Exams \cite{hardalov-etal-2020-exams}; \textbf{Commonsense reasoning tasks}  Hellaswag \cite{zellers-etal-2019-hellaswag}, PIQA \cite{Bisk2020}, SIQA \cite{sap-etal-2019-social}, BoolQ \cite{clark2019boolq}, Arc Challenge \cite{allenai:arc}, OpenBookQA \cite{OpenBookQA2018} and Winogrande \cite{DBLP:journals/corr/abs-1907-10641}; \textbf{Misinformation and Bias tasks} TruthfulQA \cite{lin2021truthfulqa} and CrowS-Pairs\cite{nangia-etal-2020-crows}.

For all models we see improvement in Arabic tasks - $7.5$\% improvement in Arabic MMLU for \textit{Llama2 70B adapted} compared to \textit{Llama2 70B} (see \ref{tab:Arabic_downstream}), while the smaller models (7B and 13B) demonstrate $2$\% improvement in MMLU Arabic. This can be attributed to the lower token-per-parameter training regime resulting in less degradation from over-training \cite{hoffmann2022training,dey2023cerebrasgpt}. We also observe  slight improvement in average scores in English for \textit{Llama2 70B adapted} (see\ref{tab:English_downstream}). Furthermore, compared to the state-of-the-art Arabic models, namely Jais and AceGPT, Arabic adapted Llama2 models significantly improve on Arabic downstream tasks.

For Arabic adaptation Llama 3 8B was continually pre-trained for 65B tokens with $1:1$ mix of English and Arabic using hyper-parameters as discussed in  \ref{hyperparameter_tuning}. In \ref{tab:Llama3_downstream_Arabic} and \ref{tab:Llama3_downstream_En} we depict the performance of \textit{Llama 3 8B adapted} on  Arabic and English tasks, respectively, and a comparison against the Llama 3 8B base model. We see $4.5$\% improvement on average in Arabic downstream tasks compared to the base model performance while a 1\% improvement in downstream English tasks. 
 
Llama 2 13B was trained on 72B tokens with $1:9$ English to Hindi dataset mixture. Learning rate was re-warmed to the pre-training peak learning rate of $3e\text{-}4$ and a batch size of 4M tokens. We evaluate on XNLI-Hindi \cite{conneau2018xnli} where the goal is to  predict textual entailment.  We see $8$\% improvement compared to Llama 2 13B while in English there is $1.8$\% degradation as shown in \ref{tab:Llama2_downstream_Hi}.

%
\section{Conclusion}
We present an efficient recipe to significantly enhance capabilities of an English-dominant foundation LLM in another language. Our approach includes extending the vocabulary, applying a novel method for embedding initialization and alignment, and continually pre-training the foundation LLM on a bilingual data mix. We perform hyper-parameter optimization for batch size, learning rate schedule, and data mix ratio to ensure successful adaptation without experiencing \quotes{catastrophic forgetting}. We successfully use this approach to enhance Arabic capability of Llama 2 base models, resulting in a state-of-the-art 70B Arabic base language model. Furthermore, we apply this approach to other languages such as Hindi and other foundation LLMs such as Llama 3 - while we briefly discuss these two studies but will leave the details for future work. We also intend to extend this approach to low resource languages and architecture such as depth-up scaling \cite{kim2024solar107bscalinglarge} and adapter layers methods as discussed in \ref{sec:Llama-pro}. 

\bibliography{main}
\bibliographystyle{icml2024}

\appendix

\section{Tokenizer ablations}
\label{tokenizer_appndx}
We experimented with one more tokenizer variant, \textit{Llama3-replace5} in addition to \textit{Llama 2-extend100} and \textit{Llama3-replace30}. Here we replace only $5$\%  of the Llama 2 tokenizaer vocabulary with that of MLV2's most frequent Arabic tokens.
Experiment design:
\begin{itemize}
    \item Continually pre-train a monolingual LLM on native + target language (Arabic) mix tokenized with the new vocabulary. 
    \item Use the same hyperparameters across the ablations. 
    \item Fix the raw text corpus for each language – this will ensure fairness as the total information/bytes are fixed.
    \item Select the size of raw text corpus for each language such that when tokenized, by a monolingual tokenizer in the respective language, the total tokens are in the same range. 
    \item \textbf{Pre-trained base model:} Llama 2-7B
    \item \textbf{Datasets:} Pile and AraV5.
    \item \textbf{Embedding initialization:} mean or subword-mean (discussed in the next section).
    \item Learning rate of $1.5e-4$ and batch size of $4$M tokens.
\end{itemize}
We take the raw corpus (bytes) of Pile and AraV5 such that when tokenized by the MLV2 Arabic tokenizer, the number of tokens for Pile and AraV5 are same.
In table \ref{tab:tokenizer_ablations_v2} we show extrinsic evaluation of different bilingual tokenizers using either \textit{extend} or \textit{replace} new token injection/merge schemes. Overall \textit{Llama 2 base} performed the best but it had $>3.5x$ more Arabic tokens compared to \textit{Llama 2-extend100} which is not compute efficient. But if we compare isoToken in terms of Arabic, \textit{Llama 2-extend100} performs the best.
 Another factor is the tokens per parameter (TPP), with \textit{Llama 2-replace5} and \textit{Llama 2-base}  the TPP is much larger, and as we’re going to be training for higher TPPs, using a high fertility tokenizer would drift far from the pareto-frontier or in other words there will be TPP degradation.
Another thing to consider here is inference cost of high fertility tokenizers as more tokens for the given text mean limited context for the model, more memory requirements.

\begin{table*}[h!]
    \centering
    \resizebox{\textwidth}{!}{\begin{tabular}{c|c c c c c c c c}
        \hline
        \hline
         \textbf{variant}     & \textbf{merge}   &  \textbf{Emb. initialization}  &  \textbf{Vocab size}   & \textbf{Total Tokens}   &   \textbf{Arabic Tokens} & \textbf{English Tokens} & \textbf{Pile loss } & \textbf{AraV5 loss }\\ \hline
        \textit{Llama 2 base}   &  N/A & N/A  & 32000 & 87.5B  & 67.31B & 20.22B &  1.508   &  0.637   \\
        \textit{Llama 2-replace5}  &  replace & Mean  & 32000 & 51B  & 30.7B  & 20.3B &  1.504   &  0.6493  \\
        \textit{Llama 2-replace5}  &  replace & Subword Mean  & 32000 & 51B  & 30.7B  & 20.3B &  1.508   &  0.646 \\
        \textit{Llama 2-replace30}  &  replace & Subword Mean  & 32000 & 43.7B  & 22.76B  & 20.94B &  1.516   &  0.6439  \\
        \textit{Llama 2-extend100}  &  extend & Mean  & 64000 & 39.7B  & 19.35B  & 20.35B &  1.499   &  0.6591  \\
        \textit{Llama 2-extend100}  &  extend & Subword Mean  & 64000 & 39.7B  & 19.35B  & 20.35B &  1.499   &  0.6539  \\
    \end{tabular}}
    \caption{Extrinsic evaluation of different bilingual tokenizers using either \textit{extend} or \textit{replace} new token injection/merge schemes. Due to varying fertility in Arabic we can see that tokenizer with smaller number of Arabic tokens in its vocabulary has a higher fertility and thus a higher token count. So this is not an iso-token comparison.}
    \label{tab:tokenizer_ablations_v2}
\end{table*}


\section{Cross Entropy Loss Correction}
\label{cec}
When comparing cross-entropy loss between models trained with data tokenized by different tokenizers, we need to apply correction or normalization to the losses. This normalization is required because cross-entropy's units of measurement are \texttt{nats/tokens}, and therefore, the definition of a token becomes very important. Depending on the size of the vocabulary and type of tokenizer, the information represented per token varies. The loss correction factor to compare cross-entropy loss between two models is then the ratio of the number of tokens in the validation sets for each model. 

\section{Embedding initialization}
\label{emb_init_appndx}
Here we discuss the ablations that we performed with different embeddings initialization methods as discussed in the main body. We also ablated an additional initialization method which we refer to as \textit{Subword Mean}. Following are the initialization methods under consideration:
\begin{itemize}
    \item \textbf{Mean}: Initialize all the new tokens’ embeddings with the mean of source language token embeddings. 

    \item \textbf{Subword Mean}: For a newly added Arabic token, tokenize it using base Llama tokenizer and use the mean of the token embeddings ot the sequence of tokens. 

    \item \textbf{Semantic similarity search based}: This method was introduced in Wechsel multilingual initialization work where a pre-trained embeddings like Fasttext or OpenAI embeddings are used. 
    
    \item \textbf{Projection based:} Use least squared to Learn a transformation matrix from a learned embedding space (MLV2) to the Llama token embeddings space using the overlapping tokens. Then apply this transformation to the newly added tokens from MLV2 vocab to Llama vocab.
\end{itemize}

Note that for LLMs with untied embeddings and unembeddings, the new tokens embeddings (or unembeddings) are initialized independently using the above methods.

In table \ref{tab:embeddings_init_ablations} we compare different initialization methods for Llama2 7B continual pre-training on AraV5 and Pile mix tokenized with \textit{Llama2-extend100}.

\begin{table*}[h!]
    \centering
    \small
    \begin{tabular}{p{3cm}|p{3cm} p{3cm} p{3cm} p{3cm} p{3cm}}
        \hline
        \hline
         \textbf{Emb. initialization}    & \textbf{Total Tokens}   &   \textbf{Arabic Tokens} & \textbf{English Tokens} & \textbf{Pile loss } & \textbf{AraV5 loss }\\ \hline
         Mean & 39.7B    & 19.35BB & 20.35B &  1.4995   &  0.6591             \\
         Subword Mean    & 39.7B  & 19.35BB & 20.35B &  1.4994   &  0.653     \\
         Wechsel $k=5$   & 39.7B  & 19.35BB & 20.35B &  1.4988   &  0.64898   \\
         Wechsel $k=10$  & 39.7B  & 19.35BB & 20.35B &  1.4999   &  0.656     \\
         MLV2 Projection & 39.7B  & 19.35BB & 20.35B &  1.5013   &  0.64857   \\
    \end{tabular}
    \caption{Comparison of different embedding initialization techniques in terms of upstream eval loss of Llama2 7B when trained on Pile and AraV5 mic for 39.7 billion tokens.}
    \label{tab:embeddings_init_ablations}
\end{table*}

\begin{table*}[h!]
    \centering
    \small
    \begin{tabular}{p{3cm}|p{4cm} p{3cm}}
        \hline
        \hline
         \textbf{Tokenizer variant} & \textbf{Emb. initialization}  & \textbf{First step train loss }\\ \hline
         \textit{Llama 2 replace5}  & Random & 16.31    \\
         \textit{Llama 2 replace5}  & Mean  & 9.15   \\
         \textit{Llama 2 extend5}   & Mean  & 9.43      \\
         \textit{Llama 2 extend100} & Mean  & 7.22     \\
         \textit{Llama 2 extend100} & Subword Mean & 5.47     \\
         \textit{Llama 2 extend100} & Wechsel $k=5$ & 5.49     \\
         \textit{Llama 2 extend100} & MLV2 projection & 4.8    \\
    \end{tabular}
    \caption{Comparison of different embedding initialization techniques across multiple tokenizers.}
    \label{tab:embeddings_init_ablations_across_tokenizers}
\end{table*}



\section{Block expansion adapter approach for multilingual models}
\label{sec:Llama-pro}
Following the work outlined in \cite{wu2024Llama} we leverage the block expansion approach for multilingual models, making it highly effective for language adaptation. By adding and fine-tuning additional Transformer blocks initialized to identity mappings, the model can integrate new domain-specific knowledge without forgetting previous information. Although, the techniques described in the original paper focus on code and math, we were able to successfully adapt the approach for our experiments with English and Arabic. We initialized our base model with Llama-2 7B and expanded the number of blocks from 32 to 40 using an interleaved approach. In our experiments for language adaptation, we found that an optimal data mix of 1:9(En:Ar) yielded the best results (in downstream 0 shot tasks in both English and Arabic) relative to adapting the newly added layers only on  domain specific data. In both experiments we trained on a total of 67B tokens in Arabic in order to maintain the same token count for the appropriate comparison. Our results show that the block-expansion approach is a strong candidate for language adaptation  with a faster time to train and lower training costs. In the future, this work could expand to other types of models(like MoE models) and modalities and would be interesting to analyse the impact on overall accuracy in downstream tasks
 For language adaptation with block expansion [section\ref{sec:Llama-pro}], we experiment with different number of adapter layers. We find that the optimal adapter layer is 25\% of the existing layer. Similarly, 960 is the optimal batch size. Table \ref{tab:block_expansion_results} summarizes our results using the above approach for language adaptation at the LLama 2 7B scale

\begin{table*}[h!]
    \centering
    \resizebox{\textwidth}{!}{\begin{tabular}{c|c c c c c c c c}
        \hline
        \hline
         \textbf{Model}     & \textbf{Data Mix(En:Ar)}   &  \textbf{Block expansion \%}  &  \textbf{Arabic tokens}   & \textbf{Arabic downstream eval acc\%}   &   \textbf{English downstream eval acc\%} \\ \hline
         
        \textit{Llama 2 7B}   &  N/A & N/A  & N/A & 38.34  & 54.68    \\
        \textit{Llama 2 7B}  &  0:1 & 25  & 67B & 42.52  & 55.69    \\
        \textit{Llama 2 7B}  &  1:9 & 25  & 67B & 43.16  & 57.80   \\        
    \end{tabular}}
    \caption{Evaluation of block expansion adapter approach 
 with data mixes across various downstream evaluation tasks. Arabic tasks are evaluated across Knowledge, Commonsense reasoning and Misinformation, bias. English tasks are evaluated for Commonsense reasoning}
    \label{tab:block_expansion_results}
\end{table*}
 
 \section{Fine-tuning}
Upstream loss is typically assumed to indicate downstream performance [\cite{isik2024scaling}, \cite{gadre2024language}]. In order to verify performance on downstream tasks in the adapted domain, we fine-tune both pre-trained and adapted pre-trained models. Instruction fine-tuning allows us to assess both performance and generation quality, which may not always match upstream performance \cite{tay2022scale}.

The data used is an extended fine-tuning dataset following from \cite{sengupta2023jais}. We add additional data in English and Arabic, focusing on bilingual examples and quality for language and style adaptation. In total, our instruction fine-tuning dataset contains approximately 10 million English and 4 million Arabic examples in both single-turn and multi-turn settings 

We fine-tuned for 3 epochs with a standard linear learning rate decay. Instead of padding, examples are packed together up to the sequence length and separated with EOS tokens, increasing training efficiency by up to 8 times. As in Jais, we calculate loss on answer tokens only.

We observe that downstream performance of the fine-tuned model trained on top of the pre-trained model is lower than that of the fine-tuned model trained after domain adaptation. It suggests that, similar to the findings in \cite{isik2024scaling}, downstream task performance after fine-tuning is highly dependent on the alignment between pre-training data and downstream tasks which is improved through adaptation.

 \section{Hardware setup}
The training runs were conducted on two Condor Galaxy supercomputers, each equipped with 64 Cerebras CS-2 Wafer-Scale Engines (WSE-2). Each CS-2 features 40 GB of SRAM and achieves a peak throughput of 7.5 PetaFLOP/s in half precision, providing a total of 960 PetaFLOP/s in half precision across both supercomputers.
Utilizing the weight streaming mode of the Cerebras software stack, the Condor Galaxy supercomputers can flexibly schedule multiple jobs based on hardware resource requirements and priority. The number of CS-2s allocated to a job can be dynamically adjusted during training, with performance scaling linearly up to 64 CS-2s per job. This scalability is facilitated by the Cerebras software stack's use of pure data parallelism to distribute the workload across multiple CS-2s. Jobs are managed by a priority queue system, ensuring efficient allocation of computational resources.

\section{Downstream Tasks}
In \ref{tab:Arabic_downstream_detailed} and \ref{tab:English_downstream_detailed} we present downstream task performance breakdown on  Arabic and English downstream tasks, respectively. We draw the comparison between Llama 2 series of Arabic adapted models compared against base Llama 2 models and open source state of the art Arabic models. \textit{Llama 2 13B}, \textit{Llama 2 13B adapted}.

Tables \ref{tab:Arabic_downstream_detailed_llama3} and \ref{tab:English_downstream_detailed_llama3} show the downstream task performance breakdown on  Arabic and English downstream tasks, respectively, for \textit{Llama 3 8B} base and \textit{Llama 3 8B adapted}.
\begin{table*}[h!]
    \centering
    \small
    \resizebox{\textwidth}{!}{\begin{tabular}{l|l l|l l l l l l|l l|l}
        ~ & Knowledge & ~ & Commonsense Reasoning & ~ & ~ & ~ & ~ & ~ & Misinformation, bias & ~ & ~ \\ \hline
        model & mmlu (acc\_norm) & exams & Hellaswag & PIQA & BoolQ(acc) & SIQA  & ARC Challenge & Openbook QA & TruthfulQA & CrowS-Pairs(pct\_stereotype) & Average \\ \hline
        llama 2-70b adapted & 37.7\% & 39.1\% & 61.6\% & 68.2\% & 66.9\% & 41.4\% & 41.2\% & 33.2\% & 45.6\% & 57.2\% & 49.2\% \\ 
        llama 2-13b adapted & 30.6\% & 37.7\% & 54.9\% & 67.1\% & 64.5\% & 40.6\% & 36.1\% & 32.0\% & 43.6\% & 54.0\% & 46.1\% \\ 
        llama 2-7b adapted & 28.7\% & 39.0\% & 48.0\% & 62.8\% & 63.9\% & 38.5\% & 32.0\% & 31.4\% & 43.9\% & 54.2\% & 43.51\% \\ \hline
        llama 2-70b base & 30.2\% & 32.6\% & 41.2\% & 54.6\% & 64.2\% & 35.2\% & 30.5\% & 31.4\% & 47.0\% & 50.9\% & 41.8\% \\ 
        llama 2-13b base & 28.4\% & 30.4\% & 34.3\% & 52.9\% & 63.8\% & 36.4\% & 24.3\% & 30.0\% & 45.5\% & 49.9\% & 39.6\% \\ 
        llama 2-7b base & 27.8\% & 26.7\% & 32.3\% & 50.0\% & 63.8\% & 35.6\% & 25.0\% & 29.0\% & 46.7\% & 48.3\% & 38.5\% \\ \hline
        Acegpt-13b & 29.9\% & 34.7\% & 45.6\% & 60.3\% & 63.2\% & 38.1\% & 32.8\% & 32.2\% & 45.1\% & 56.4\% & 43.8\% \\ \hline
        Jais 30b & 34.0\% & 42.0\% & 60.4\% & 69.0\% & 67.7\% & 42.2\% & 39.2\% & 33.8\% & 45.1\% & 57.3\% & 49.1\% \\
    \end{tabular}}
    \caption{Detailed break down of \textbf{Knowledge}, \textbf{Commonsense Reasoning} and \textbf{Misinformation, bias} downstream Arabic task performance of Llama 2 adapted models, and comparison against base Llama 2 and state of the art Arabic open source models}
    \label{tab:Arabic_downstream_detailed}
\end{table*}

\begin{table*}[h!]
    \centering
    \small
    \resizebox{\textwidth}{!}{\begin{tabular}{l|l l|l l l l l l l|l l|l}
       ~ & Knowledge & ~ & Commonsense Reasoning & ~ & ~ & ~ & ~ & ~ & ~ & Misinformation, Bias & ~ & ~ \\ \hline
        model & mmlu & race & Hellaswag & PIQA & BoolQ & SIQA  & ARC Challenge & Openbook QA & Winogrande & TruthfulQA & CrowS-Pairs(pct\_stereotype) & Average \\ \hline
        llama 2-70b adapted & 52.2\% & 39.6\% & 82.0\% & 81.5\% & 82.8\% & 46.2\% & 52.1\% & 45.8\% & 75.9\% & 43.8\% & 68.2\% & 60.9\% \\ 
        llama 2-13b adapted & 37.3\% & 39.5\% & 76.5\% & 78.6\% & 77.8\% & 44.6\% & 45.9\% & 44.4\% & 71.4\% & 34.6\% & 64.0\% & 55.9\% \\ 
        llama 2-7b adapted & 33.9\% & 38.6\% & 74.0\% & ~ & 75.4\% & 44.4\% & 42.2\% & 43.6\% & 67.3\% & 37.6\% & 65.7\% & 52.3\% \\ \hline
        llama 2-70b base & 55.6\% & 42.0\% & 80.8\% & 81.0\% & 76.8\% & 42.6\% & 48.0\% & 44.4\% & 76.9\% & 44.5\% & 69.6\% & 60.2\% \\ 
        llama 2-13b base & 34.9\% & 40.8\% & 76.6\% & 79.1\% & 69.0\% & 44.9\% & 44.3\% & 42.0\% & 69.6\% & 37.6\% & 69.8\% & 55.3\% \\ 
        llama 2-7b base & 32.0\% & 40.1\% & 73.0\% & 77.0\% & 71.1\% & 42.7\% & 40.5\% & 40.8\% & 67.2\% & 39.6\% & 71.1\% & 54.1\% \\ \hline
        Acegpt-13b & 34.6\% & 39.7\% & 77.0\% & 79.6\% & 77.6\% & 45.7\% & 44.2\% & 40.0\% & 70.1\% & 39.4\% & 73.7\% & 56.5\% \\ \hline
        Jais 30b & 42.3\% & 40.3\% & 79.1\% & 80.5\% & 80.9\% & 49.3\% & 48.4\% & 43.2\% & 70.6\% & 40.3\% & 72.3\% & 58.8\% \\ 
    \end{tabular}}
    \caption{Detailed break down of \textbf{Knowledge}, \textbf{Commonsense Reasoning} and \textbf{Misinformation, bias} downstream English task performance of Llama 2 adapted models, and comparison against base Llama 2 and state of the art Arabic open source models}
    \label{tab:English_downstream_detailed}
\end{table*}

\begin{table*}[h!]
    \centering
    \small
    \resizebox{\textwidth}{!}{\begin{tabular}{l|l l|l l l l l l|l l|l}
        ~ & Knowledge & ~ & Commonsense Reasoning & ~ & ~ & ~ & ~ & ~ & Misinformation, bias & ~ & ~ \\ \hline
        model & mmlu (acc\_norm) & exams & Hellaswag & PIQA & BoolQ(acc) & SIQA  & ARC Challenge & Openbook QA & TruthfulQA & CrowS-Pairs(pct\_stereotype) & Average \\ \hline
        llama 3-8b & 31.0\% & 34.7\% & 45.2\% & 57.7\% & 66.4\% & 37.5\% & 34.6\% & 30.0\% & 48.3\% & 46.4\% & 43.2\% \\ 
        llama 3-8b adapted & 36.0\% & 41.5\% & 51.8\% & 64.5\% & 71.4\% & 41.2\% & 35.8\% & 31.0\% & 52.1\% & 54.4\% & 47.9\% \\ 
    \end{tabular}}
    \caption{Detailed break down of \textbf{Knowledge}, \textbf{Commonsense Reasoning} and \textbf{Misinformation, bias} downstream Arabic task performance of Llama 3 8B adapted model, and comparison against base Llama 3 8B}
    \label{tab:Arabic_downstream_detailed_llama3}
\end{table*}

\begin{table*}[h!]
    \centering
    \small
    \resizebox{\textwidth}{!}{\begin{tabular}{l|l l|l l l l l l l|l l|l}
       ~ & Knowledge & ~ & Commonsense Reasoning & ~ & ~ & ~ & ~ & ~ & ~ & Misinformation, Bias & ~ & ~ \\ \hline
        model & mmlu & race & Hellaswag & PIQA & BoolQ & SIQA  & ARC Challenge & Openbook QA & Winogrande & TruthfulQA & CrowS-Pairs(pct\_stereotype) & Average \\ \hline
        llama 3-8b & 62.0\% & 40.1\% & 79.2\% & 80.9\% & 81.2\% & 33.1\% & 53.4\% & 45.0\% & 72.8\% & 43.9\% & 62.3\% & 59.5\% \\ 
        llama 3-8b adapted & 60.6\% & 39.0\% & 79.1\% & 80.5\% & 82.3\% & 33.2\% & 53.5\% & 46.2\% & 73.3\% & 51.1\% & 66.7\% & 60.5\% \\ 
    \end{tabular}}
    \caption{Detailed break down of \textbf{Knowledge}, \textbf{Commonsense Reasoning} and \textbf{Misinformation, bias} downstream English task performance of Llama 3 8B adapted model, and comparison against base Llama 3 8B}
    \label{tab:English_downstream_detailed_llama3}
\end{table*}

\end{document}